# FACE-net: Factual Calibration and Emotion Augmentation for Retrieval-enhanced Emotional Video Captioning

Weidong Chen, *Member, IEEE*, Cheng Ye, Zhendong Mao, *Senior Member, IEEE*, Peipei Song, Xinyan Liu, Lei Zhang, Xiaojun Chang, *Senior Member, IEEE*, Yongdong Zhang, *Fellow, IEEE*

**Abstract**—Emotional Video Captioning (EVC) is an emerging task, which aims to describe factual content with the intrinsic emotions expressed in videos. Existing works perceive global emotional cues and then combine with video content to generate descriptions. However, insufficient factual and emotional cues mining and coordination during generation make their methods difficult to deal with the factual-emotional bias, which refers to the factual and emotional requirements being different in different samples on generation. To this end, we propose a retrieval-enhanced framework with **FA**ctual **C**alibration and **E**motion augmentation (FACE-net), which through a unified architecture collaboratively mines factual-emotional semantics and provides adaptive and accurate guidance for generation, breaking through the compromising tendency of factual-emotional descriptions in all sample learning. Technically, we firstly introduces an external repository and retrieves the most relevant sentences with the video content to augment the semantic information. Subsequently, our factual calibration via uncertainty estimation module splits the retrieved information into subject-predicate-object triplets, and self-refines and cross-refines different components through video content to effectively mine the factual semantics; while our progressive visual emotion augmentation module leverages the calibrated factual semantics as experts, interacts with the video content and emotion dictionary to generate visual queries and candidate emotions, and then aggregates them to adaptively augment emotions to each factual semantics. Moreover, to alleviate the factual-emotional bias, we design a dynamic bias adjustment routing module to predict and adjust the degree of bias of a sample. Extensive experiments, especially remarkable improvements on the CIDEr and CFS metrics, on three challenging datasets demonstrate the superiority of our approach and each proposed module.

**Index Terms**—Emotional Video Captioning, Retrieval Augmentation, Multimodal Emotion Understanding.

✦

## 1 INTRODUCTION

WITH the development of online technology, more and more people are beginning to express their opinions on social platforms by videos, which can arouse people's emotional resonance due to its vividness and variability. Under such circumstances, video emotional analysis tasks have attracted more and more attention, including video emotion classification [1], [2], [3], video advertising recommendation [4], [5], [6], and emotional video captioning [7], [8], [9]. Among them, emotional video captioning (EVC) has been widely researched by the community nowadays due to the complexity of its multimodal properties, which requires not only understanding the factual video contents, but also recognizing the complex emotion cues contained in the video, and then incorporating the emotion semantics to generate captions. EVC has wide applications in real-world scenarios, such as social media video analysis or advertising recommendations.

EVC has dual requirements for factual and emotional factors. Existing methods [8], [9], [10], [11] perceive global emotions at first, and then combine them with video content to generate factual descriptions with emotions. However, based on our observations of video data, we find that samples are inevitably biased towards fact or emotion. As illustrated in Fig. 1, some samples tend to describe emotions (*emotional bias*), some samples tend to describe facts (*factual bias*), and others tend to treat both fact and emotion important (*neutral samples*).[1] Existing methods ignore above characteristic. Overall learning across all samples causes the model to approach the compromised results of emotional and factual descriptions, leading to factual-emotional compromised homogenized results. As shown in Fig. 1, when handling factual or emotional bias samples, the most powerful state-of-the-art method DCGN [9] generates compromised results. Two insufficient considerations make previous methods fail to handle the factual-emotional bias. Firstly, their methods usually solve factual and emotional modeling issues in isolation, which oversimplifies the complex factual semantics and the obscure emotional semantics modeling without considering the synergy between facts

This research is supported by Artificial Intelligence-National Science and Technology Major Project 2023ZD0121200, and the National Natural Science Foundation of China under Grants 62121002, 62302474, 62402471, and 62502115. Corresponding author: Zhendong Mao.

Weidong Chen, Cheng Ye, Peipei Song, Lei Zhang, and Xiaojun Chang are with the School of Information Science and Technology, University of Science and Technology of China, Hefei 230027, China (e-mail: chenweidong@ustc.edu.cn; kyrieye@mail.ustc.edu.cn; beta.songpp@gmail.com; leizh23@ustc.edu.cn; xjchang@ustc.edu.cn).

Xinyan Liu is with the School of Computer Science and Technology, Harbin Institute of Technology (Weihai), Weihai, China (e-mail: xinyliu@hit.edu.cn).

Zhendong Mao and Yongdong Zhang, are with the School of Information Science and Technology, University of Science and Technology of China, and are also with the Institute of Artificial Intelligence, Hefei Comprehensive National Science Center, Hefei 230027, China (e-mail: zdmao@ustc.edu.cn; zhyd73@ustc.edu.cn).

1. The statistical verification of our observation is shown in Sec. 3.1.

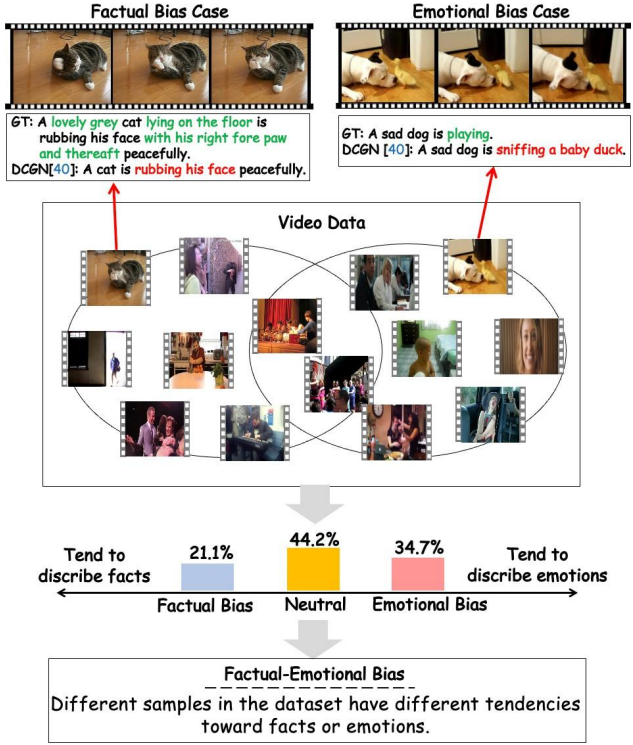

Fig. 1. Motivation of our method, including data bias statistics and performance analysis of the existing method.

and emotions, resulting in insufficient semantics mining. Besides, nonadaptive coordination between factual and emotional semantics for different samples makes the model generate captions to be biased toward the compromised results. Therefore, we regard it as the most important issue in our paper and name it as **factual-emotional bias**, which is that *different samples in the dataset have different tendencies toward facts or emotions.*

To address the limitation, we propose a novel retrieval-enhanced framework with **FA**ctual **C**alibration and **E**motion augmentation (FACE-net) for emotional video captioning. Different from previous methods, our method adopts a unified architecture to first calibrate retrieved facts and then augment emotions. Its joint approach integrates the synergy of facts and emotions to achieve accurate semantics mining. Besides, our method adaptively augments emotions for facts as well as collaborates all factual-emotional information by a dynamic routing, providing adaptive guidance for caption generation, which breaks through the inherent factual-emotional requirements of samples and the descriptive tendency of all samples in compromise learning. Its adaptability combined with accurate semantics mining can generate personalized captions for different samples, thus alleviating the factual-emotional bias problem.

Technically, we first retrieve sentences related to the video from an external repository by cosine similarity ranking to enhance semantic information richness. Meanwhile, we segment sentences into subject-predicate-object triples for pure and efficient representations. Subsequently, to remove redundant information of triples, we propose a Factual Calibration via Uncertainty Estimation module. Specifically, we utilize entropy theory to quantify the redundant information content in triples, thereby refining the individual components of the triples. Next, we assign weights to each triple by calculating the mutual information between the video and each triple, which eliminates the influence of irrelevant triple information and achieves accurate fact semantic mining.

In addition, to accurately capture visual emotional cues and provide emotional guidance for caption generation, we propose a Progressive Visual Emotion Augmentation module. Specifically, we regard factual semantics as query experts and design a progressive emotion mining paradigm. First, we leverage factual semantics to extract potential emotion candidates from an open emotion dictionary. Secondly, we combine the video content with factual semantics to generate visual queries. Finally, we leverage visual queries to capture authentic emotional semantics from emotional candidates. Such a progressive mining approach improves the stability and accuracy of emotion capturing.

To quantify and eliminate the degree of the factual-emotional bias in video samples, we propose a Dynamic Bias Adjustment Routing module. Specifically, we first design a bias adjustment component with a gated unit, which accurately quantifies the biases of emotion and fact by learning different video samples. Subsequently, we design a dynamic routing component to calculate the contributions of each retrieval group to the bias adjustment. Finally, we leverage route weights to aggregate all factual and emotional semantics, along with their corresponding biases, as multimodal input to a LLM-based decoder for emotional caption generation. Extensive experiments on three challenge datasets demonstrate the effectiveness of our method and each proposed module. In short, our main contributions are summarized as follows:

· We first observe that different samples in the data set have different tendencies towards facts or emotions, and define the **factual-emotional bias** problem through statistical analysis. To alleviate the factual-emotional bias, we propose a retrieval-enhanced network with **FA**ctual **C**alibration and **E**motion augmentation (FACE-net) for the EVC task, which retrieves triplet information from external repositories to collaboratively mine factual-emotional semantics and provide adaptive and accurate guidance for generation, breaking through the compromising tendency of factual-emotional descriptions in overall learning.

· Technically, we propose a factual calibration via uncertainty estimation module to mine factual semantics, which quantifies the triple uncertainty through entropy theory to achieve feature refinement and remove redundant information. Meanwhile, we propose a progressive visual emotion augmentation module to capture emotional semantics, which serves factual semantics as experts, interacts with video contents and emotion dictionary to obtain visual queries and candidate emotions, and then aggregates them to augment visual emotions.

· Extensive experiments on three public benchmarks (*e.g.*, EVC-MSVD, EVC-VE, and EVC-Combine) [7] prove the effectiveness of our method and each proposed module. Especially on CIDEr/CFS metrics, which emphasizes semantic relevance, our model reaches almost doubled performance of 169.4/153.6 on EVC-MSVD.

## 2 RELATED WORK

In this section, we briefly investigate existing methods, referring to retrieval-enhanced methods, multimodal emotional analysis, and emotional video captioning.

### 2.1 Retrieval-enhanced Methods

retrieval-enhanced techniques have been proven to be effective components for many deep learning tasks, which leverage external knowledge to boost the performance [12], [13], [14], [15]. Especially in the multimodal vision-language domain, researchers retrieve semantically related samples to enhance tasks, such as image recognition [16], [17], video captioning [18], [19], and knowledge-based visual question answering [20], [21]. They typically leverage the powerful cross-modal alignment capabilities of the CLIP model to enhance retrieval-augmentation. In the captioning task, Sarto et al. [22] introduce an external kNN memory with a knowledge retriever that utilizes visual similarities to improve generation quality. Xu et al. [23] develop a system that automatically retrieves semantically relevant third-person instructional videos to enhance caption generation. Li et al. [24] introduce EVCap, which constructs a dynamic object knowledge memory using external visual object names to facilitate open-world understanding. In the visual question answering task, Chen et al. [25] access an external non-parametric multimodal memory to augment language generation, which is pretrained with a mixture of large-scale imagetext and text-only corpora using a joint contrastive and generative loss. Li et al. [26] focus on internet-augmented generation (IAG), which provides up-to-date knowledge via internet search during inference, and propose the SearchLVLMs, which trains a hierarchical filtering model to effectively and efficiently find the most helpful content from the websites returned by a search engine. Chen et al. [27] propose the RagVL to address the limitation that the existing models rely on static training data, leading to outdated information and limited contextual awareness, particularly in dynamic or rapidly evolving contexts.

However, existing approaches face significant challenges when applied to emotion-related tasks. The primary limitations are twofold. 1) Irrelevant emotional content in retrieved information can adversely affect model performance. 2) Effectively integrating factual and emotional semantics from retrieved information remains challenging. To address these limitations, we propose our FACE-net framework in this work to effectively utilize retrieval-enhanced techniques to improve the quality of emotional descriptions.

### 2.2 Visual Emotional Analysis

Instead of solely focusing on the factual content, understanding the emotional cues contained in visual elements is increasingly becoming a popular topic in community research [28], [29], [30], [31], [32]. Due to the variability and ambiguity of emotions, it is an extreme challenge to accurately capture them in visual content. Some researchers regard the visual emotional analysis as a common classification problem and propose the emotional classification or recognition task. They typically design models based on traditional convolutional neural networks or transformer architectures and leverage pre-defined emotion sets to assign labels to images/videos. Li et al. [33] propose a hierarchical CNN-RNN approach to predict the emotion based on the fused stimuli by exploiting the dependency among different-level features. specifically, they leverage a dual-CNN to extract different levels of visual stimulus and a stacked bi-directional RNN to fuse the preceding learned features from the dual CNN. Inspired by the Stimuli-OrganismResponse (S-O-R) emotion model in psychological theory, Yang et al. [34] propose a stimuli-aware VEA method, containing stimuli selection (S), feature extraction (O), and emotion prediction (R). Zhu et al. [35] propose an image-text interaction network to explore the relationship between image affective regions and the associated text. Specifically, they introduce a cross-modal alignment module to capture region-word correspondence, so that multimodal features are fused through an adaptive cross-modal gating module. Subsequently, they integrate the individual-modal contextual feature representations for achieving more reliable predictions. Besides, some researchers are dedicated to exploring the causes and interpretability of visual emotions. Zhang et al. [36] introduce and fuse the valence-arousal-dominance (VAD) knowledge to better understand and generate emotional labels and their explanation texts. Panos et al. [37] collect a large scale dataset containing images, emotional responses, and free-form text interpretation, and develop a novel model to provide plausible affective responses to real-world visual data explained with language.

Despite promising progress, we still face some challenges when performing the visual emotional analysis in EVC task. On the one hand, due to the high redundancy of long videos, capturing emotion-related visual elements to generate accurate emotional representations is extremely challenging. Besides, we are puzzled about what kind of emotional representation could effectively guide the generation of emotional descriptions. To address these limitations, we design the progressive visual emotion augmentation module to capture accurate visual emotional semantics.

### 2.3 Emotional Video Captioning

Emotional Video Captioning (EVC) extends beyond traditional captioning by incorporating emotional cues into generated captions. Traditional methods [38], [39], [40], [41] have relied on pre-defined emotion categories (i.e., happiness, surprise, disgust, anger) to produce emotionally-aware captions, but this approach often overlooks the implicit emotional cues present in video content. Wang et al. [7] pioneer this field by introducing a visual emotion analyzer to detect implicit emotional content, along with a dual-stream network that integrates emotional and factual semantics for caption generation. They also contributed a comprehensive EVC dataset to the research community. Subsequently, Song et al. [8] propose a contextual attention network to recognize and describe the fact and emotion in the video by semantic-rich context learning. Besides, Song et al. [10] develop a tree-structured emotion repository enabling hierarchical emotion perception. Building upon this work, Song et al. [11] design a multimodal context fusion module that combines visual, textual, and visual-textual relevance to enhance EVC performance. Wang et al. [42] introduce two learnable prompting strategies: visual emotion prompting and textual emotion prompting, to learn emotional cue representations and further design two levels of objective functions: the ER-sentence level and the AU-word level alignment losses, to facilitate the interaction and alignment. Ye *et al.* [43] focus on the importance of emotional causes in emotional exploration and propose a multi-round mutual learning network to jointly extract emotion-cause pairs, which enhances the accuracy of emotion mining and the interpretability of emotional descriptions.

While these approaches have advanced the field, they struggle to integrate factual and emotional semantics in all-sample learning and fail to handle the factual-emotional bias problem, which is the emphasis of research in our paper. Ye *et al.* [9] propose a novel framework to leverage fact-emotion collaborative learning to address the challenge of dynamic emotion changes during generation. However, it is considered a local collaborative approach, which only coordinates intra-sample emotions and facts at each generation step. Thus, in general, they cannot overcome the different inter-sample needs of facts and emotions, and compromise the overall learning results.

## 3 METHOD

The core goal of EVC task is to generate a factually and sentimentally accurate sentence to describe the given video. As mentioned above, we observe the factual-emotional bias issue (F-E bias) in EVC task. In Sec. 3.1, we make a statistical analysis to verify the existence. Subsequently, we propose a factual calibration and emotion augmentation (FACE-net) framework to solve it. As shown in Fig. 2, our proposed FACE-net mainly consists of five parts: *Retrieval-based Multimodal Feature Extraction* (Sec. 3.2), *Factual Calibration via Uncertainty Estimation* (Sec. 3.3), *Progressive Visual Emotion Augmentation* (Sec. 3.4), *Dynamic Bias Adjustment Routing* (Sec. 3.5). The main research questions are three: 1) how to model factual semantics to generate content-correct descriptions (Q1), 2) how to effectively capture visual emotional cues and provide accurate emotional guidance for caption generation (Q2), and 3) how to alleviate the factual-emotional bias problem (Q3).

### 3.1 Preliminary Statistical Analysis for F-E Bias

Preliminarily, we first observe that samples in EVC datasets are inevitably biased towards fact or emotion, and define it as factual-emotional bias (F-E bias). To verify our observation, we make a detailed statistical analysis of F-E bias on the public benchmark EVC-MSVD [7] with a careful consideration. Two reasonable criteria for statistics are defined:

· **Rule 1:** Firstly, we believe that the proportion of the emotional descriptions in the entire caption could reflect the sample bias. Thus, we set two thresholds, *i.e.*, $t_1$ =1/6 and $t_2$ =1/10, to distinguish between factual bias, neutral, and emotional bias. Specifically, samples with an emotional description ratio higher than $t_1$ (*at least one emotional word in $1/t_1$-word or less sentence*) and less than $t_2$ (*at most one emotional word in $1/t_2$-word or more sentence*) are considered to be emotional bias and factual bias, respectively.

· **Rule 2:** On the basis of Rule 1, we further analyze the neutral and factual bias samples by comparing with MSVD

dataset, which is homologous dataset to EVC-MSVD. EVC-MSVD is further annotated based on MSVD, and both two datasets have common video data. Since MSVD is a dataset for traditional video captioning task, which does not require the annotator to consider emotional factors when labeling. Therefore, for a video sample, if the caption annotation of MSVD describes emotional captions, we believe that this sample biased toward emotional bias.

Based on Rule 1, we count the proportions of emotional bias, neutral, and factual bias as 29.0%, 47.4%, and 23.6%, respectively. Furthermore, with the refinement of Rule 2, the results after Rule 1 are further justified. We efficiently count the number of each category, which are 34.7%, 44.2%, and 21.1%, respectively, and intuitively demonstrate the existence of F-E bias.

### 3.2 Feature Extraction and Retrieval Enhancement

**Video Encoder.** We first extract the sample-frame sequence $\{f_1, f_2, \ldots, f_N\}$ from the video, where $N$ is the frame number of the video. Following [44], to enhance temporal feature extraction, we leverage a 3D convolution with kernel $[t \times h \times w]$ as the linear instead of the kernel of $[h \times w]$ in 2D linear, where $t$, $h$, and $w$ are temporal, height, and width dimensions, respectively. Finally, we exploit the transformer to model the interaction between each patch of the input image to get the final features $V = \{v_1, v_2, \ldots, v_N\} \in \mathbb{R}^{N \times d_v}$. The above process is formalized as:

$$V = \text{VideoEncoder}(\phi_{3D}(\{f_1, f_2, \ldots, f_N\})), \quad (1)$$

where $\phi_{3D}$ denotes the 3D linear projection.

**Emotion Encoder.** Following previous work [10], [11], we leverage a open psychology emotional vocabulary dictionary $D_w = \{e_i\}_{i=1}^{N_w}$ where $e_i$ denotes the $i$-th emotion word, such as "happy", and $N_w$ denotes the number of emotion words. Then, we use the pre-trained embedding model GloVe [45] to encode these emotion words to $E_0 \in \mathbb{R}^{N_w \times d_E}$. The process can be formalized as:

$$E_0 = \text{TextEncoder}(D_w). \quad (2)$$

It is worth noting that $E_0$ is the embedding of the overall emotion dictionary, which is used to augment emotions to each factual semantics.

For the issue (Q1), it is crucial for the accuracy and logic of descriptions to mine accurate factual semantics. Inspired by the rapid development of RAG technology, we attempt to implement retrieval-enhanced factual content mining through external corpus repositories. Specifically, we collect a textual repository with a large number of descriptions $C = \{c_i\}_{i=1}^{N_c}$ where $c_i$ is a sentence with rich factual semantic information, including objects, actions, and attributes. Subsequently, we implement video-text retrieval by CLIP [46], a pre-trained model that is considered to have strong cross-modal alignment capabilities. With the common embedding space of CLIP, we encode the visual content and textual repository as follows:

$$\hat{v} = \text{MeanPool}(\text{CLIP}_V(\{f_i\}_{i=1}^{N})), \quad (3)$$

$$w_i = \text{CLIP}_T(c_i), \quad (4)$$

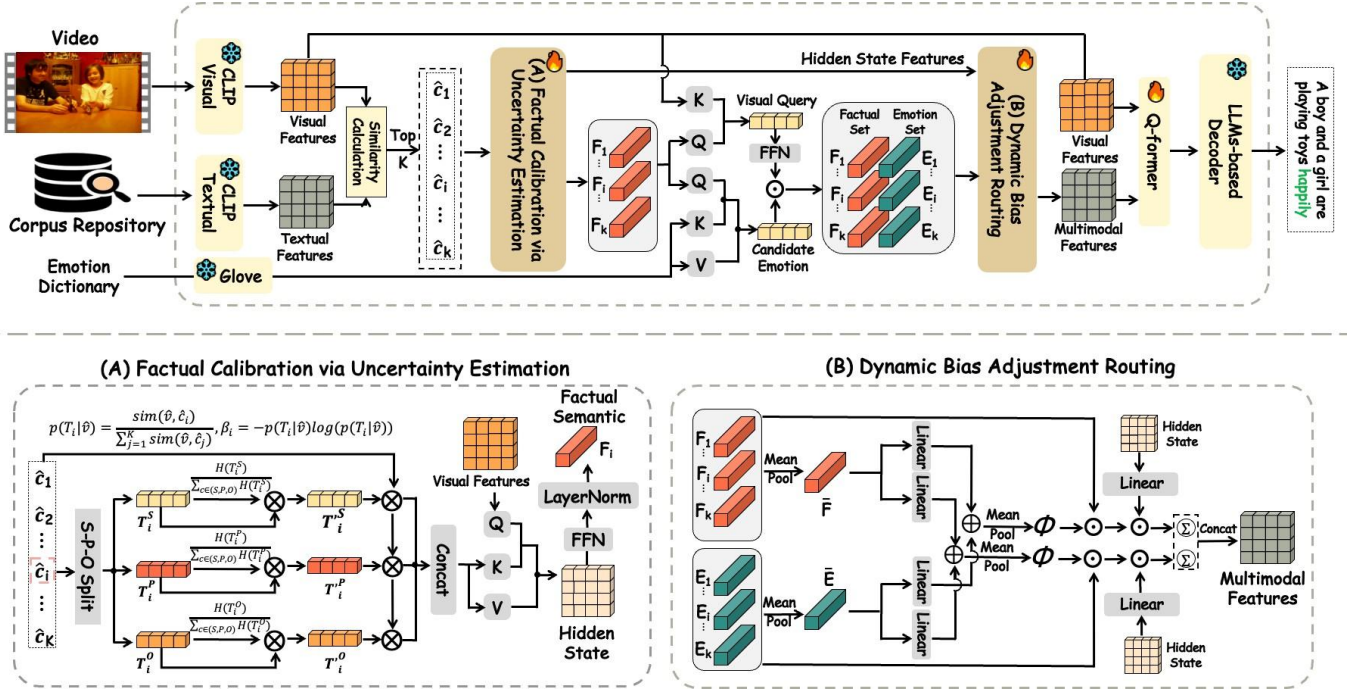

Fig. 2. The proposed FACE-net consists of three main components: Factual Calibration via Uncertainty Estimation (FCUE), Progressive Visual Emotion Augmentation (PVEA), and Dynamic Bias Adjustment Routing (DBAR). Our model firstly retrieves the most relevant sentences to the video content from an external corpus repository to provide rich auxiliary semantics for caption generation. Subsequently, these sentences are fed into the FCUE module, which calibrates video-irrelevant noise through self-refinement and cross-refinement to generate factual semantics. Then, the PVEA module utilizes factual semantics and a pre-defined psychological emotion dictionary to mine fine-grained emotion semantics from video content. The DBAR module adjusts the factual-emotional bias through a gated unit component and assigns the weights of each triplet through a mixture-of-experts component to obtain the multimodal representations. Finally, we aggregate the multimodal representations through a lightweight Q-former component and feed them into LLMs to generate factually and sentimentally accurate descriptions.

where MeanPool denotes the mean-pooling operation to aggregate the features of all frames into compact visual representations. We rank all sentences by cosine similarity between visual representation and each textual representation, and finally select the top-$K$ relevant sentences $C = \{c^{\hat{}}_i\}_{i=1}^K$. The above process could be formulated as:

$$sim(v^{\hat{}}, w_i) = \frac{w^{\hat{}}_i{}^\top v^{\hat{}}}{\|w_i\|\|v\|}, \tag{5}$$

$$C^{\hat{}} = C_{\underset{i}{argmax}\ sim(v^{\hat{}}, w_i)}, \tag{6}$$

### 3.3 Factual Calibration via Uncertainty Estimation

However, we observe that this simple similarity-based retrieval method is difficult to significantly improve the quality of generated descriptions. Similarity calculations are performed on the global feature dimension and are difficult to filter out noise information irrelevant to the video in a fine-grained manner. Therefore, the sub-problem of issue (**Q1**) arises: **how to eliminate noise from retrieval results to obtain concise and efficient auxiliary semantic information (Q1.1).**

Firstly, instead of leveraging the whole sentence, we choose to extract the subject-predicate-object (S-P-O) triplet of the sentence, which is considered to be the condensation of the core semantics and removes semantic-irrelevant textual noise. Specifically, we use the SOTA relation extraction model [47] to extract the triplets $\{(\mathsf{S}_i, \mathsf{P}_i, \mathsf{O}_i)\}$, where

$i \in [1, K]$, and S, P, O denote subject, predicate, and object, respectively. Subsequently, we leverage CLIP$_\mathsf{T}$ to extract the triplet embedding $T_i$ by adding a unified prefix H such as "A photo of" on the triplet $(\mathsf{S}_i, \mathsf{P}_i, \mathsf{O}_i)$:

$$(\mathsf{S}_i, \mathsf{P}_i, \mathsf{O}_i) = \mathsf{RelationModel}(c^{\hat{}}_i), \tag{7}$$

$$T_i = \mathsf{CLIP}_\mathsf{T}(\mathsf{H} + (\mathsf{S}_i, \mathsf{P}_i, \mathsf{O}_i)) \in \mathsf{R}^{3 \times d_T}, \tag{8}$$

$$T = \{T_1, T_2, \ldots, T_K\} \in \mathsf{R}^{K \times 3 \times d_T}, \tag{9}$$

Moreover, each triplet contains different semantic uncertainty for factual semantic mining. Therefore, we design a two-step semantic refinement module, which refines triple features by quantifying uncertainty and obtains more accurate factual semantics. Specifically, we first perform self-refinement on each triple itself. We leverage information entropy theory to calculate the semantic gain of the three components $\{(\mathsf{S}, \mathsf{P}, \mathsf{O})\}$:

$$\mathbf{p}^c_i = \mathsf{Softmax}(T^c_i),\ c \in \{\mathsf{S}, \mathsf{P}, \mathsf{O}\}, \tag{10}$$

$$H(T^c_i) = -\sum \mathbf{p}^c log(\mathbf{p}^c), \tag{11}$$

$$\alpha^c_i = \frac{H(T^c_i)}{\sum_{c \in \{\mathsf{S}, \mathsf{P}, \mathsf{O}\}} H(T^c_i)}, \tag{12}$$

$$T' = \boldsymbol{\alpha} \odot T \in \mathsf{R}^{K \times 3 \times d_T}, \tag{13}$$

where $\odot$ denotes the element-wise multiplication operation. Afterwards, to reduce redundancy and avoid inflated entropy, we perform cross-refinement among all triplets. Specifically, we first compute the aggregated probability of each triplet $(S_i, P_i, O_i)$ as $p(T_i)$, reflecting its share of the total similarity mass. Then we perform cross-refinement by the semantic entropy operation as follows:

$$p(T_i | v^{\wedge}) = \sum_i \frac{sim(v^{\wedge}, w_i)}{\sum_{j=1}^{K} sim(v^{\wedge}, w_j)}, \quad (14)$$

$$\beta_i = \sum_i \frac{p(T_i | v^{\wedge}) log(p(T_i | v^{\wedge}))}{\sum_{j=1}^{K} p(T_j | v^{\wedge}) log(p(T_j | v^{\wedge}))}, \quad (15)$$

$$\overline{T} = \beta \odot T' \in \mathbb{R}^{K \times 3 \times d_T}, \quad (16)$$

where $\odot$ denotes the element-wise multiplication operation. So far, we have obtained the calibrated triple features with rich and accurate auxiliary semantic information. Finally, we fuse triple features with video content to capture factual semantics:

$$H_i = \text{Attn}_f(V, [\overline{T_i}W_T])|_{Q:V,\{K,V\}:[\overline{T_i}W_T]}, \quad (17)$$

$$F_i = \text{LayerNorm}(\text{FFN}([V; H_i]) + V), \quad (18)$$

where $W_T \in \mathbb{R}^{d_T \times d_V}$ is a learnable weight to project triple features into the same embedding space as $V$, and $[;]$ is a concatenate operation. The hidden features $H_i \in \mathbb{R}^{N \times d_V}$ not only achieve alignment with the video factual content, but also serve as an expert weight for retrieved information aggregation. Meanwhile, $F_i \in \mathbb{R}^{N \times d_V}$ contains the pure and accurate factual semantics, playing an important role in content-correct descriptions and could serve as a crucial cornerstone for visual emotion cues mining.

### 3.4 Progressive Visual Emotion Augmentation

Mining accurate visual emotional cues plays a crucial role in generating emotion-related words and enhancing the vividness and readability of descriptions. Therefore, in this section, we focus on resolving (Q2) based on two inferences. Firstly, we believe that the key emotional triggers should be focused on the main factual content of the video (*i.e.*, objects and actions), which is precisely summarized by the factual semantics. Therefore, we first leverage the calibrated factual semantics to mine candidate emotions cues from the pre-defined emotion dictionary:

$$\tilde{E}_i = \text{Attn}_c(F_i, E_0)|_{Q:F_i,\{K,V\}:E_0} \in \mathbb{R}^{N \times d_E}. \quad (19)$$

Secondly, a triplet could express a variety of different emotions in different contexts. For example, the emotion is usually sadness for triplets (*girl, lose, teeth*). However, losing baby teeth means growing up, which may make girls happy. Therefore, we combine video features and factual semantics to generate visual queries, which contain contextual information about the video content. Finally, we leverage visual queries to further filter candidate emotions for accurate emotional semantics:

$$q_v = \text{Softmax}(\phi_F(F_i) \cdot \phi_V(V^\top)) \in \mathbb{R}^{N \times N}, \quad (20)$$

$$E_i = q_v \cdot \phi_E(\tilde{E}_i) \in \mathbb{R}^{N \times d_E}, \quad (21)$$

where $\phi$ denotes the linear projection. Through a two-stage progressive emotion mining process, we obtain emotional semantics $E_i$ for $i$-th retrieval group, which not only matches the triplet information of the current retrieval group, but also adapts to the video context, ensuring accurate emotion guidance for description generation.

### 3.5 Dynamic Bias Adjustment Routing

After obtaining sufficient factual semantic $F_i$ and emotional semantic $E_i$ for each triplet, to effectively alleviate the factual-emotional bias problem (Q3), we design a dynamic bias adjustment routing module. Firstly, we design a bias adjustment component by a gated unit to balance the weights of factual and emotional semantics:

$$\overline{F} = \text{MeanPool}([F_1, F_2, \dots, F_K]), \quad (22)$$

$$\overline{E} = \text{MeanPool}([E_1, E_2, \dots, E_K]), \quad (23)$$

$$B_f = \phi(\text{MeanPool}(FU_f + \overline{E}R_f + b_f)), \quad (24)$$

$$B_e = \phi(\text{MeanPool}(FU_e + \overline{E}R_e + b_e)), \quad (25)$$

where $U_f, R_f, U_e, R_e \in \mathbb{R}^{d \times d}$, $b_f, b_e \in \mathbb{R}^d$, MeanPool and $\phi$ denote the average pooling and *tanh* function, respectively. $B_f, B_e$ generalize the bias by learning factual and emotional representations from different video samples, generating accurate gating thresholds to alleviate the F-E bias problem. Moreover, to quantify the contributions of each retrieval group, we design a dynamic routing component to assign weights to each retrieval group. Specifically, all hidden features $H_i$ are concatenated and then fed into a dynamic router, which is a linear transformation of the input followed by a softmax layer:

$$G = \text{Softmax}([H_1; H_2; \dots; H_k]W_g), \quad (26)$$

where $W_g$ is a learnable parameter. The output $G = \{g_1, g_2, \dots, g_K\}$ preserves the predicted weights for diverse triplets. $g_i$ is the elements of $G$ for $\forall i$. Finally, we calculate the aggregated multi-modal representation as follows:

$$M = [B_f \sum_{i=0}^{K} g_i F_i; B_e \sum_{i=0}^{K} g_i E_i] \in \mathbb{R}^{2N \times d}, \quad (27)$$

where $d = d_E = d_V$ is the common dimension in our work. The multi-modal representation $M$ contains sufficient factual and emotional semantics of retrieved captions, which serve as auxiliary semantical cues for caption generation.

With sufficient multimodal semantics, we utilize LLMs to generate captions due to their powerful representation and generation capabilities. However, for the EVC task, LLMs are difficult to directly understand previously generated multimodal representations. Thus, we leverage Q-former in BLIP-2 [48], [49], [50], [51] to convert them to tokens that could be interpreted by LLMs. Specifically, the Q-former uses learnable query vectors $q \in \mathbb{R}^{N_q \times d}$ to extract interpretable representations, where $N_q$ is the query length. We first use self-attention to aggregate each query:

$$\overline{q} = \text{Attn}_q(q)|_{\{Q,K,V\}:q} + q \in \mathbb{R}^{N_q \times d}, \quad (28)$$

then the aggregated query is applied to generate interpretable multimodal representations by a cross-attention layer and a linear projection:

$$\overline{M} = \phi_M(\text{Attn}_d(\overline{q}, [M;V])|_{Q:q,\{K,V\}:[M;V]} + \overline{q}) \in \mathbb{R}^{N_q \times d}, \quad (29)$$

Finally, we send $\overline{M}$ to the LLMs to generate emotional descriptions. Since the important role of emotional words

TABLE 1
Performance for emotional video captioning task on three benchmarks. The best results are highlighted in bold. The second best results are marked by underline. The percentage in blue indicates how much our method improves over the second best method.

| Methods | Emotion | | Semantic | | | | | | | Hybrid | |
|---|---|---|---|---|---|---|---|---|---|---|---|
| | Acc$_{sw}$↑ | Acc$_c$↑ | BLEU-1↑ | BLEU-2↑ | BLEU-3↑ | BLEU-4↑ | METEOR↑ | ROUGE-L↑ | CIDEr↑ | BFS↑ | CFS↑ |
| **EVC-MSVD** | | | | | | | | | | | |
| SA-LSTM [52] | 68.8 | 67.2 | 80.7 | 67.9 | 56.3 | 45.5 | 33.0 | 68.2 | 72.1 | 59.0 | 71.3 |
| FT [7] | 69.4 | 67.1 | 77.2 | 60.3 | 47.4 | 36.3 | 29.0 | 63.4 | 62.5 | 52.5 | 63.7 |
| CANet [8] | 78.7 | 76.8 | 78.5 | 64.0 | 52.1 | 41.8 | 30.8 | 65.7 | 74.4 | 57.9 | 75.1 |
| VEIN [11] | 82.7 | 82.1 | 82.0 | 68.4 | 57.1 | 45.9 | 33.0 | 69.0 | 79.6 | 62.4 | 80.2 |
| EPAN [10] | 84.1 | 82.8 | 82.5 | 69.6 | 57.8 | 46.2 | 34.4 | 69.8 | 80.6 | 63.1 | 81.1 |
| DCGN [9] | 86.5 | 85.7 | 84.5 | 70.9 | 59.2 | 48.7 | 35.7 | 71.0 | 85.2 | 65.7 | 86.6 |
| VideoBLIP [53] | 85.7 | 84.7 | 94.1 | 85.7 | 77.5 | 69.7 | 43.4 | 81.6 | 159.4 | 78.8 | 142.6 |
| MM-ECPE [43] | <u>90.4</u> | <u>89.1</u> | <u>96.9</u> | <u>87.8</u> | <u>80.2</u> | <u>71.4</u> | <u>45.5</u> | <u>83.1</u> | <u>168.3</u> | <u>81.8</u> | <u>152.6</u> |
| **FACE-net(Ours)** | **91.3** | **89.6** | **97.5** | **88.2** | **80.5** | **71.5** | **46.1** | **83.7** | **169.4** | **82.2** | **153.6** |
| **EVC-VE** | | | | | | | | | | | |
| SA-LSTM [52] | 48.6 | 47.1 | 71.0 | 51.1 | 34.5 | 22.5 | 19.6 | 40.7 | 30.2 | 38.9 | 33.7 |
| CANet [8] | 41.9 | 39.7 | 66.9 | 44.8 | 29.3 | 19.3 | 18.2 | 37.9 | 23.3 | 33.9 | 26.8 |
| VEIN [11] | 57.4 | 56.8 | 71.6 | 52.1 | 37.4 | 26.3 | 20.9 | 41.7 | 33.4 | 43.0 | 39.2 |
| EPAN [10] | 63.8 | 62.3 | 73.6 | 54.0 | 38.3 | 27.0 | 21.2 | 42.3 | 34.7 | 45.0 | 40.4 |
| DCGN [9] | 71.0 | 69.4 | 74.5 | 55.3 | 40.0 | 28.1 | 23.4 | 47.7 | 41.5 | 47.3 | 46.9 |
| VideoBLIP [53] | 59.6 | 58.8 | 74.4 | 55.3 | 39.7 | 27.4 | 22.8 | 48.2 | 61.2 | 46.8 | 60.9 |
| MM-ECPE [43] | <u>73.4</u> | <u>72.3</u> | <u>76.8</u> | <u>57.5</u> | <u>41.7</u> | <u>28.9</u> | **24.7** | <u>49.5</u> | <u>65.2</u> | <u>49.2</u> | <u>66.0</u> |
| **FACE-net(Ours)** | **73.9** | **72.6** | **77.5** | **58.1** | **42.2** | **29.3** | <u>24.5</u> | **49.7** | **66.4** | **49.7** | **67.8** |
| **EVC-Combine** | | | | | | | | | | | |
| SA-LSTM [52] | 53.4 | 50.7 | 70.6 | 51.4 | 36.7 | 25.4 | 21.0 | 45.9 | 38.8 | 41.2 | 41.5 |
| FT [7] | 51.2 | 49.6 | 67.6 | 47.2 | 32.0 | 21.6 | 20.4 | 43.1 | 29.0 | 37.6 | 33.3 |
| CANet [8] | 53.7 | 52.7 | 68.1 | 47.7 | 32.9 | 22.5 | 19.7 | 43.7 | 34.5 | 38.8 | 38.2 |
| VEIN [11] | 59.0 | 57.6 | 72.1 | 52.8 | 37.9 | 27.1 | 21.6 | 46.8 | 39.4 | 43.6 | 43.1 |
| EPAN [10] | 69.3 | 67.2 | 74.4 | 55.6 | 39.9 | 28.0 | 23.0 | 47.1 | 43.0 | 47.0 | 48.0 |
| DCGN [9] | 74.8 | 73.1 | 75.6 | 56.7 | 40.5 | 28.5 | 24.9 | 51.7 | 49.8 | 48.5 | 51.7 |
| VideoBLIP [53] | 58.4 | 57.2 | 74.1 | 56.0 | 40.8 | 29.2 | 25.3 | 52.0 | 65.6 | 48.5 | 63.6 |
| MM-ECPE [43] | <u>75.6</u> | <u>73.8</u> | <u>78.1</u> | <u>58.5</u> | <u>42.3</u> | <u>30.2</u> | <u>26.4</u> | <u>53.8</u> | <u>67.9</u> | <u>50.4</u> | <u>69.3</u> |
| **FACE-net(Ours)** | **75.9** | **74.4** | **78.7** | **59.1** | **43.4** | **30.9** | **26.9** | **54.0** | **69.2** | **51.1** | **70.5** |

in EVC, we use an emotion-focused cross-entropy loss that adds a penalty term on emotional words:

$$L_e = \begin{cases} -(1+\delta)\sum_{N_T}^{t=1} \log P(y_t|y_{<t}), & \text{if } y_t \in D_w, \\ -\sum_{N_T}^{t=1} \log P(y_t|y_{<t}), & \text{otherwise.} \end{cases} \quad (30)$$

where $\delta$ is a hyper-parameter that controls the level of punishment when $y_t$ is an emotional word like "happily", and $N_T$ is the sequence length. Furthermore, to provide a more substantial emotional loss, we also design an emotional classification loss. Specifically, we add a simple classification header for the weighted sum $E$ of augmented emotion $E_i$ of each triple to obtain the emotional distribution $d$. Then we calculate the emotional classification loss:

$$L_{cls} = -\sum_{e \in E} \log P(e|\overline{D_w}). \quad (31)$$

The overall loss of our method is the weighted sum of these two losses:

$$L = \lambda_e L_e + \lambda_{cls} L_{cls}, \quad (32)$$

where $\lambda_e, \lambda_{cls}$ are two hyper-parameters that aim to control the balance of two losses. In doing so, our model can be end-to-endly trained by minimizing the overall loss $L$.

# 4 EXPERIMENT

## 4.1 Experimental Settings

**Datasets.** We test our model on three datasets. **EVC-MSVD** [7] is built on the basis of MSVD [54], which has 240/134

videos, 8169/4611 sentences for training/testing, respectively. **EVC-VE** [7] is built based on VE-8 [55], which contains 1141/382 videos and 19398/6527 sentences for training/testing, respectively. **EVC-Combine** [7] is the combination of EVC-MSVD and EVC-VE.

**Metrics.** To quantify the performance of captions generated by our model, we leverage metrics including BLEU 1-4 [56], METEOR [57], ROUGE-L [58], and CIDEr [59] to evaluate the factual semantics. To measure the accuracy of emotional words and emotional sentences, we leverage two emotional accuracy metrics Acc$_{sw}$ and Acc$_c$ [7] and two hybrid metrics BFS and CFS, which combine BLEU and CIDEr with emotional accuracy metrics.

**Implementation Details.** For each video, we sample 16 frames and resize them to 224 × 224 with central cropping. We extract video features with CLIP [46]. We build an overall vocabulary that contains all words in the corpus [10]. The vocabulary sizes for the EVC-MSVD, EVC-VE, and EVC-Combine [7] datasets are 9,637, 13,980, and 14,034, respectively. The number of retrieved captions is set to $K = 4$ from a repository, which is composed of captions in the training dataset except the current input video. We extract the triplet for each retrieved caption by UniRel [47]. The number of emotion words is $N_w = 179$, following [60], [61]. All embedding dimensions are unified to $d = d_V = d_T = d_E = 300$. For the Q-former component, we set the length of the learnable queries to 32. During training, we freeze the LLM decoder and integrate a LoRA adapter [62] with $r = 16$ and $\alpha = 32$ to align the video and emotional latent space with the LLM decoder. We adopt the Adam optimizer [63] with a learning rate of 7e-4, and the batch size is set to 32. We set the penalty parameter $\delta$, the objective function weight $\lambda_e$ and $\lambda_{cls}$ to 0.1/0.2/0.1, 1.0/1.0/1.0, and 0.1/0.5/0.2 for EVC-MSVD/EVC-VE/EVC-Combine dataset, respectively. The maximum length of captions is set to 15 and the size of beam search is set to 5. All experiments are implemented on 2 NVIDIA A800 GPUs.

TABLE 2
The results of ablation studies on EVC-Combine dataset for different decoders.

| Methods | Acc$_{SW}$↑ | Acc$_C$↑ | BLEU-1↑ | BLEU-2↑ | BLEU-3↑ | BLEU-4↑ | METEOR↑ | ROUGE-L↑ | CIDEr↑ | BFS↑ | CFS↑ |
|---|---|---|---|---|---|---|---|---|---|---|---|
| **Classical Decoder** | | | | | | | | | | | |
| EPAN [10]$_{LSTM}$ | 69.3 | 67.2 | 74.4 | 55.6 | 39.9 | 28.0 | 23.0 | 47.1 | 43.0 | 47.0 | 48.0 |
| DCGN [9]$_{LSTM}$ | 74.8 | 73.1 | 75.6 | 56.7 | 40.5 | 28.5 | 24.9 | 51.7 | 49.8 | 48.5 | 51.7 |
| **FACE-net(Ours)**$_{LSTM}$ | 75.4 | 74.1 | 76.8 | 57.8 | 41.7 | 29.6 | 25.5 | 52.3 | 61.1 | 48.8 | 60.9 |
| **FACE-net(Ours)**$_{Transformer}$ | 75.1 | 73.8 | 76.5 | 57.8 | 41.2 | 28.9 | 25.1 | 52.6 | 60.4 | 48.5 | 60.2 |
| **LLMs-based Decoder** | | | | | | | | | | | |
| VideoBLIP [53]$_{FLAN-T5\ XL}$ | 58.4 | 57.2 | 74.1 | 56.0 | 40.8 | 29.2 | 25.3 | 52.0 | 65.6 | 45.8 | 63.6 |
| MM-ECPE [43]$_{FLAN-T5\ XL}$ | 75.6 | 73.8 | 78.1 | 58.5 | 42.3 | 30.2 | 26.4 | 53.8 | 67.9 | 50.4 | 69.3 |
| **FACE-net(Ours)**$_{OPT-2.7B}$ | 75.2 | 73.4 | 78.2 | 58.8 | 42.8 | 30.6 | 26.8 | 53.5 | 67.4 | 50.6 | 68.8 |
| **FACE-net(Ours)**$_{FLAN-T5\ XL}$ | **75.9** | **74.4** | **78.7** | **59.1** | **43.4** | **30.9** | **26.9** | **54.0** | **69.2** | **51.1** | **70.5** |

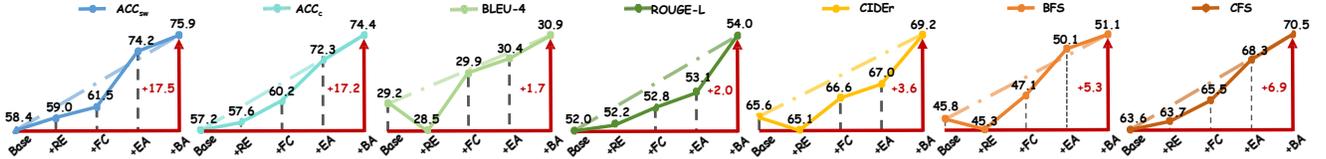

Fig. 3. The results of ablation studies on EVC-Combine dataset to discuss the effectiveness of our proposed components. RE, FC, EA, and BA stand for video-text retrieval via semantic similarity, factual calibration via uncertainty estimation, progressive visual emotion augmentation, and dynamic bias adjustment routing module, respectively.

**Baselines.** We consider several state-of-the-art methods to make comparisons and divide them into two categories: (i) Traditional Captioning Methods and (ii) Emotional Captioning Methods.

*(i) Traditional Captioning Methods*

1) **SA-LSTM** (CVPR18') [52] proposes a reconstruction network to leverage both the forward (video to sentence) and backward (sentence to video) flows for video captioning. The forward flow produces the sentence description based on the encoded video semantic features and the backward flow reproduces the video features based on the hidden state sequence generated by the decoder.

2) **CANet** (TMM22') [8] proposes a contextual attention network to recognize and describe the fact and emotion by semantic-rich context learning, which first extracts visual and textual features from both video and previously generated words and then applies the attention mechanism to capture informative contexts for captioning.

3) **VideoBLIP** (EMNLP24') [53] proposes a training paradigm that induces in-context learning over video and text by capturing key properties of pre-training data found by prior work to be crucial for improving the ability of LLMs to generate video descriptions.

*(ii) Emotional Captioning Methods*

1) **FT** (TMM21') [7] introduces a visual emotion analyzer to detect implicit emotional cues and a dual-stream network

that integrates emotional and factual semantics for caption generation by a weighted sum operation.

2) **EPAN** (MM23') [10] introduces a tree-structured emotion repository to enable hierarchical emotion perception and an emotion-prior awareness network to achieve the explicit and fine-grained emotion perception by an emotion masking mechanism and then decode the emotional caption by exploiting the multimodal semantic cues.

3) **VEIN** (TIP24') [11] designs a vision-based emotion interpretation network, which first models the emotion distribution over an open psychological vocabulary and then incorporates visual context, textual context, and visual-textual relevance into an aggregated multimodal contextual vector to enhance video captioning.

4) **DCGN** (MM24') [9] propose a framework to leverage fact-emotion collaborative learning to address the challenge of dynamic emotion changes during caption generation.

5) **MM-ECPE** (MM25') [43] focus on the importance of emotional causes in emotional exploration and propose a multi-round mutual learning network to jointly extract emotion-cause pairs for LLM-based caption generation.

### 4.2 Comparison with State-of-the-art Methods

As shown in Table 1, we present the comparison with previous methods on three datasets. We evaluate the performance of our model from three types of metrics, *i.e.*, accuracy, semantic, and hybrid metrics. Our proposed model outperforms the state-of-the-art methods for most metrics.

Firstly, we test our model on semantic metrics, which could reflect the accuracy and vividness of generated captions. We observe that our model significantly surpasses existing methods on most semantic metrics. For instance, on the EVC-MSVD dataset, our model performs better than DCGN by 15.3%/17.9% on BLEU-1/ROUGE-L metrics, respectively. Especially on the CIDEr metric, our model achieves a huge improvement of 98.8%. Due to the collaborative modeling and adaptive coordination between factual calibration and emotion augmentation module, our model could mine accurate semantics and provide customized guidance for caption generation, which generates more semantically relevant descriptions.

TABLE 3
The results of ablation studies on the EVC-Combine dataset for different numbers of retrieved captions.

| top-$K$ | Acc$_{sw}$↑ | Acc↑ | BLEU-1↑ | BLEU-2↑ | BLEU-3↑ | BLEU-4↑ | METEOR↑ | ROUGE-L↑ | CIDEr↑ | BFS↑ | CFS↑ |
|---------|-------------|------|---------|---------|---------|---------|---------|----------|--------|------|------|
| 1 | 73.8 | 71.5 | 73.4 | 55.2 | 39.7 | 28.3 | 24.6 | 50.9 | 63.9 | 47.8 | 65.7 |
| 2 | 74.2 | 72.0 | 75.4 | 56.6 | 40.9 | 28.8 | 25.1 | 51.7 | 64.5 | 48.7 | 66.2 |
| 3 | 75.1 | 73.1 | 76.3 | 58.2 | 42.4 | 29.7 | 25.9 | 52.9 | 65.6 | 49.9 | 67.3 |
| 4 | **75.9** | **74.4** | **78.7** | **59.1** | **43.4** | **30.9** | **26.9** | **54.0** | **69.2** | **51.1** | **70.5** |
| 5 | 74.8 | 72.8 | 75.8 | 57.6 | 41.5 | 29.3 | 25.4 | 52.1 | 65.0 | 49.4 | 66.8 |

TABLE 4
The results of ablation studies on EVC-Combine dataset to discuss the effectiveness of each dense component in factual calibration module. SR and CR stand for self-refinement and cross-refinement, respectively.

| Components | | A$_{sw}$↑ | A$_c$ ↑ | B4↑ | M↑ | C↑ | BFS↑ | CFS↑ |
|---|---|---------|--------|-----|-----|-----|------|------|
| SR | CR | | | | | | | |
| × | × | 72.4 | 71.6 | 29.1 | 25.9 | 65.9 | 48.6 | 67.1 |
| ✓ | × | 73.0 | 72.4 | 29.3 | 26.2 | 66.4 | 49.1 | 67.7 |
| × | ✓ | 74.3 | 73.5 | 29.9 | 26.5 | 66.9 | 49.7 | 68.3 |
| ✓ | ✓ | **75.9** | **74.4** | **30.9** | **26.9** | **69.2** | **51.1** | **70.5** |

TABLE 5
The comparisons of different orders. E-first and F-first stand for emotion augmentation-first and factual calibration-first, respectively.

| Order | A$_{sw}$↑ | A$_c$↑ | B4↑ | M↑ | C↑ | BFS↑ | CFS↑ |
|-------|---------|-------|-----|-----|-----|------|------|
| E-first | 71.5 | 70.7 | 28.9 | 25.4 | 63.8 | 48.6 | 65.3 |
| F-first | **75.9** | **74.4** | **30.9** | **26.9** | **69.2** | **51.1** | **70.5** |

Secondly, it is an extremely important purpose to predict accurate emotion for the EVC task. Thus, we evaluate the emotional accuracy of our model. It can be observed that our model achieves the best performances on emotion accuracy metrics across three datasets, i.e., 4.1%/4.6% improvements on EVC-VE over DCGN [9] with Acc$_{sw}$/Acc$_c$ metrics. The proposed model FACE-net utilizes calibrated factual semantic cues to augment the emotion mining process, which enhances the emotional understanding capabilities and predicts the emotions more accurately.

Furthermore, we consider combining the accuracy with BLEU and CIDEr metrics respectively to investigate the overall performance of our model. As shown in the last two columns of Table 1, our model achieves the best performance with BFS and CFS metrics on three datasets, i.e., 1.6%/1.7% improvements over MM-ECPE [43] on EVC-VE/EVC-Combine datasets with CFS metric, which demonstrates that captions generated by our model take into account both emotional and factual accuracy.

### 4.3 Ablation Study

To further demonstrate the superiority of FACE-net, we conduct ablation studies from four perspectives for discussion. Firstly, we discuss the effectiveness of each proposed module and dense components on the factual calibration module. Secondly, we compare the performance of our model with SOTA methods under different decoder settings to eliminate unfair comparisons caused by differences in decoders. Meanwhile, we discuss the differences between the fact-first and emotion-first framework. Finally, we discuss the impact of several key hyper-parameters in our work.

**Discussion on each component.** We perform an ablation study on each component of our model. As shown in Fig.

5, we observe that using the RE module alone brings only slight improvement for our model, and even reduces the performance on some metrics, i.e., 2.4%/0.8% decreases on BLEU-4/CIDEr metrics. We analyze that it is due to the negative impact of irrelevant noise in retrieved information. Besides, our factual calibration module, which leverages video content to calibrate factual semantics, further improves the performance, i.e., 4.0%/2.8% improvements on BFS/CFS metrics. To further improve the ability to describe emotions, we introduce the emotion augmentation module to achieve accurate emotion cues mining, which brings 20.7%/20.1% improvements on Acc$_c$/Acc$_{sw}$ metrics. Finally, we introduce the dynamic bias adjustment module to collaborate the factual and emotional semantics to alleviate the F-E bias problem, which brings 1.7%/3.3% improvements on ROUGE-L/CIDEr metrics.

**Discussion on factual calibration.** Factual calibration is one of the most important modules of our model. Thus, we also conduct a discussion on dense components consisting of self-refinement (SR) and cross-refinement (CR). As shown in Table 4, solely leveraging SR component could only bring slight improvement, i.e., 0.8%/0.7% on Acc$_{sw}$/BLEU-4. The SR component only quantifies the semantic information within each triplet, ignoring its relevance to the visual content, thus struggling to eliminating noise essentially. Instead, the CR component calculates the mutual information between each triplet and the video content, effectively filtering out the most relevant triplets to eliminate noise, resulting in a significant improvement, i.e., 2.7%/2.3% on Acc$_c$/METEOR. Finally, we combine these two components together to bring a remarkable performance, i.e., 4.8%/5.1% improvement on Acc$_{sw}$/BFS.

**Discussion on different decoders.** We discuss the generation performance of different decoders for fair comparisons, including two classical decoders LSTM [64] and Transformer [65], and two LLMs-based decoders OPT-2.7B [66] and FLAN-T5 XL [67]. As shown in Table 2, we observe that even using the same decoder with previous works, i.e., LSTM as DCGN [9], our model still performs better, i.e., 2.4%/1.2% improvements with METEOR/ROUGE-L metrics on EVC-Combine datasets. Meanwhile, we observe that although VideoBLIP [53] leverages an LLM-based decoder and achieves good performance in NLG (semantic) metrics, it is difficult to capture accurate emotional semantics, resulting in poor performance in emotional caption generation, especially in terms of emotional accuracy metrics and hybrid metrics. For example, our model leads VideoBLIP on $Acc_{sw}$ and $Acc_c$ by 30.0% and 30.1%. Besides, although the most powerful EVC method MM-ECPE [43] also conducts emotional modeling, it ignores the factual-emotional bias problem and performs holistic learning between samples, causing the compromised results of learning emotional and factual descriptions. Instead, our model efficiently coordinates factual and emotional semantics through factual calibration and emotion augmentation modules, thereby alleviating the factual-emotional bias problem. Thus, our model performs better than MM-ECPE in all metrics, *i.e.*, 1.4%/1.7% improvements with BFS/CFS metrics on EVC-Combine datasets. Finally, for two different LLMs-based decoder, we observe that FLAN-T5 XL performs better than OPT-2.7B, which is analyzed that FLAN-T5 XL has larger parameters and stronger generalization ability than OPT-2.7B, making it more suitable for the EVC task.

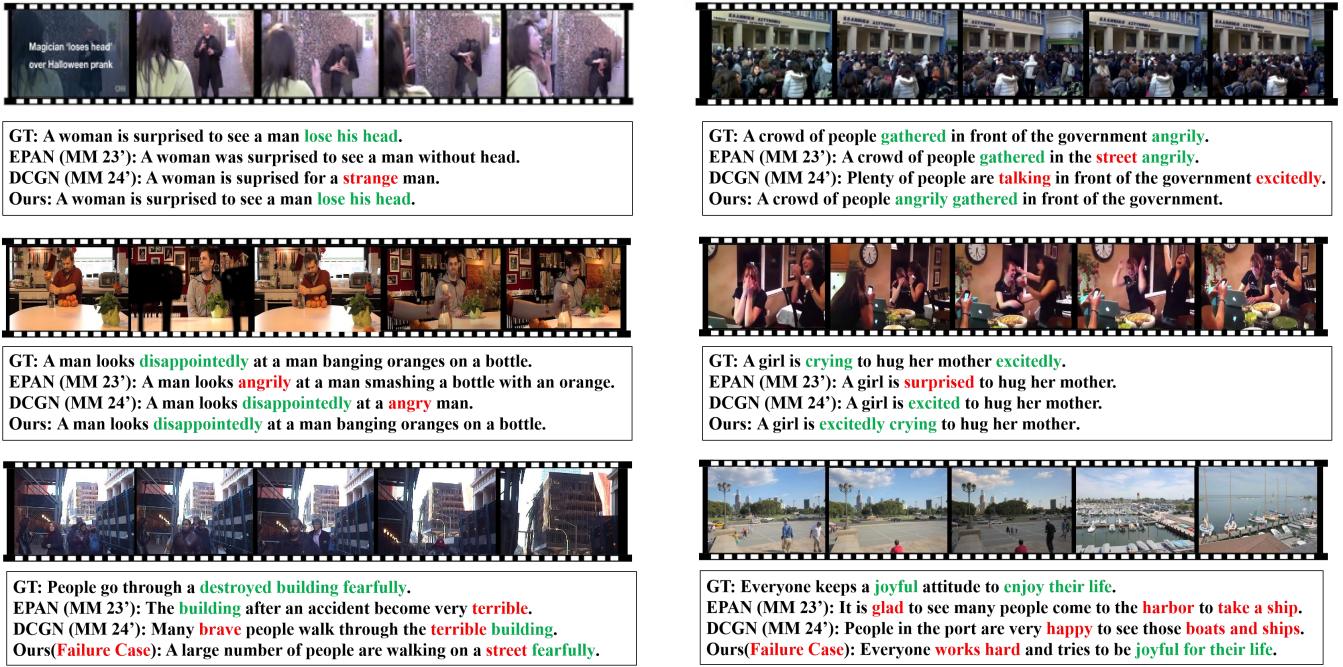

Fig. 4. Qualitative results on EVC-VE dataset to make a visualized comparison between our model and the state-of-the-art methods, *i.e.*, EPAN [10] and DCGN [9].

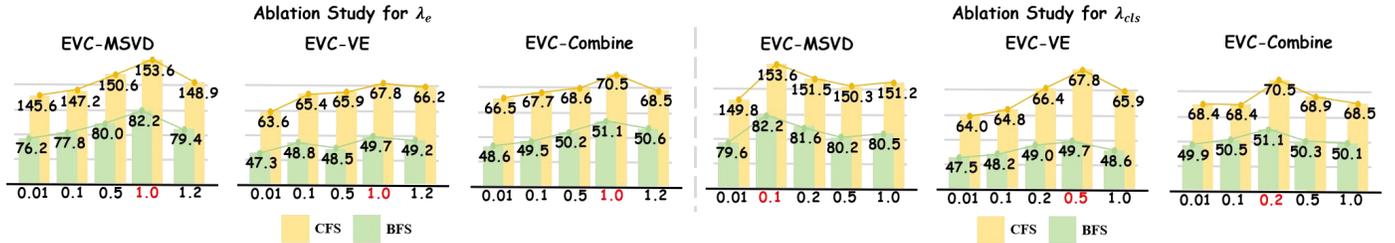

Fig. 5. The results of the ablation study for different hyper-parameters $\lambda_e$ and $\lambda_{cls}$, which aims to control the balance of emotion-focused cross-entropy loss and emotional classification loss.

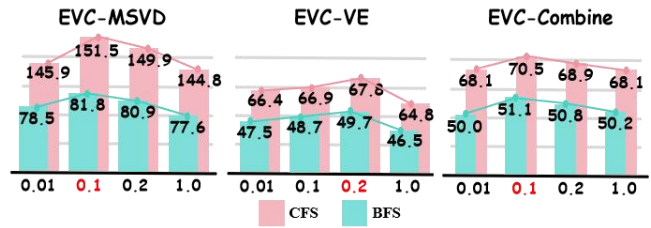

Fig. 6. The results of the ablation study for different hyper-parameters $\delta$, which aims to controls the level of punishment for emotion-related words.

**Discussion on fact-first framework.** The order of mining factual and emotional semantics is also one of the important factors. As shown in Table 5, the performance of the fact-first framework is significantly advanced, *i.e.,* 6.2% and 8.0% performance improvements on $Acc_{sw}$ and CFS, respectively. Similar to how humans perceive emotions, the accurate emotional exploration should be based on the correctness of the factual content, which fully demonstrates the rationality of our factual calibration-first framework.

**Discussion on retrieval number.** The number of retrieved captions is a significant hyper-parameter for our model.

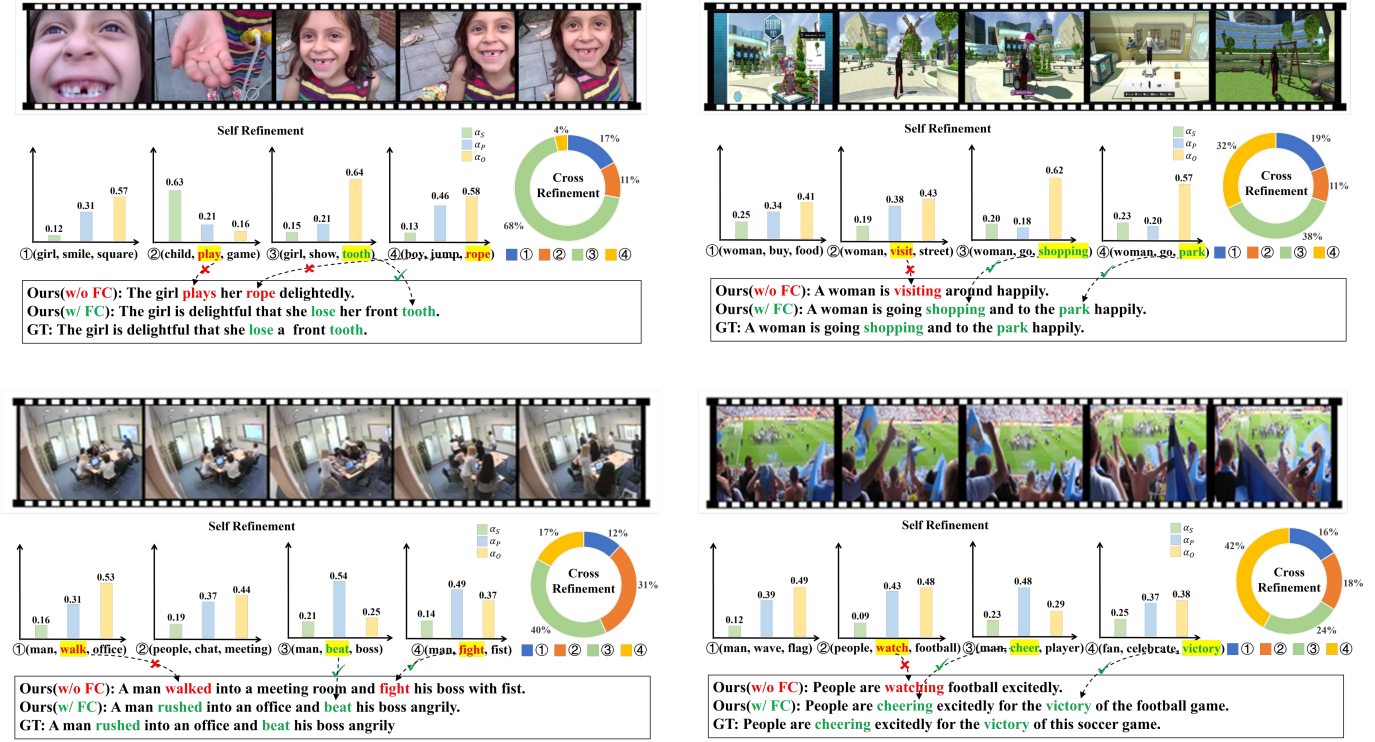

Fig. 7. Qualitative results to demonstrate the effectiveness of our proposed factual calibration module. The bar chart shows the normalized weights of the words within each triplet. The pie chart shows the normalized weights of each triplet.

TABLE 6
Comparison with the state-of-the-art methods on the efficiency and complexity evaluation, including the number of trainable parameters, the average inference time for a single caption, and the amount of memory during the inference stage.

| Method | CFS | Parameters | FLOPs | Execution Times | Memory |
|---|---|---|---|---|---|
| EPAN [10] | 48.0 | 21.1M | 0.72G | 198.6ms | 5602MB |
| DCGN [9] | 51.7 | 21.9M | 1.24G | 247.8ms | 6201MB |
| VideoBLIP [53] | 63.6 | 53.7M | 3.31G | 3610.0ms | 15852MB |
| MM-ECPE [43] | 69.3 | 60.6M | 3.72G | 4076.8ms | 18742MB |
| FACE-net(Ours) | 70.5 | 63.9M | 3.98G | 4252.6ms | 21064MB |

TABLE 7
The results of human evaluations.

| Metric | Accuracy | Relevance | Coherence | Usability |
|---|---|---|---|---|
| VEIN [11] | 5.47 | 5.18 | 6.27 | 5.04 |
| EPAN [10] | 6.24 | 5.97 | 7.13 | 5.76 |
| DCGN [9] | 6.88 | 6.43 | 7.68 | 6.32 |
| VideoBLIP [53] | 5.91 | 5.54 | 7.20 | 4.56 |
| MM-ECPE [43] | 6.40 | 6.75 | 8.04 | 7.36 |
| FACE-net(Ours) | **7.62** | **7.24** | **8.50** | **8.06** |

Thus, we conduct an ablation study to discuss the performances of different values of top-$K$. As illustrated in Table 3, we observe that our model achieves the best performance with $K = 4$. On the one hand, too small retrieval samples are hard to provide sufficient semantic information, but instead amplify the negative impact of irrelevant information. On the other hand, too much retrieval information may affect the model to mine semantic cues of the video content itself. Therefore, the above experiments show that we choose an appropriate $K$ value for our model.

**Discussion on $\lambda_e$ and $\lambda_{cls}$.** In this paper, we leverage two kinds of objective functions to jointly optimize our model. It is important to choose proper hyper-parameters $\lambda_e$ and $\lambda_{cls}$ to balance each objective function. Thus, we conduct an ablation study in this section to discuss the best setting of $\lambda_e$ and $\lambda_{cls}$, and the results are shown in Fig. 5. It can be seen that our model achieves the best performance with $\lambda_e = 1.0/1.0/1.0$ and $\lambda_{cls} = 0.1/0.5/0.2$ for for three datasets, respectively. $\lambda_e$ controls the emotional-related cross-entropy loss function $L_e$, which introduces an additional emotion punishment on emotion words, forcing the model to generate correct words at each time step. It is the reason that we should assign a main weight to it. $\lambda_{cls}$ controls the emotional classification loss $L_{cls}$ calculated by emotional distribution. Therefore, considering the overall performance of our model, we choose a balanced hyper-parameter setting of $\lambda_e$ and $\lambda_{cls}$.

**Discussion on $\delta$.** Different from the conventional cross-entropy loss, $\delta$ is an important hyper-parameter in the emotion-focused cross-entropy loss that controls the level of punishment when $y_t$ is an emotional word. Thus, we conduct an ablation study to discuss the best choice of $\delta$. As shown in Fig. 6, we observe that our model achieves the best performance with $\delta = 0.1/0.2/0.1$ for three datasets, respectively. A proper punishment coefficient could balance the dual generation requirements for factual and emotional words and generate high-quality captions.

### 4.4 Efficiency and Complexity Evaluation

Besides, the efficiency and computational complexity are crucial metrics for real-world applications. Thus, we make a comparison with SOTA methods on four efficiency metrics: parameter, FLOPs, execution time, and memory size. Firstly, compared with models that leverage traditional decoders (LSTM or Transformer), even though the overhead of the model inevitably increases, we significantly improve the model's performance while keeping the inference cost of the model to a single NVIDIA 3090 GPU (21064MB). Compared with EPAN and DCGN, our model improves the CFS metric by 46.9% and 36.4%, respectively. Secondly, compared with models that also leverage the LLM-based decoder, our model significantly improves performance with only a slight increase in computational cost. For instance, compared with VideoBLIP [53], our model achieves a 10.8% performance improvement with only an additional 5212MB of video memory, which demonstrates the comprehensive capability of our model on both metric performance and efficiency.

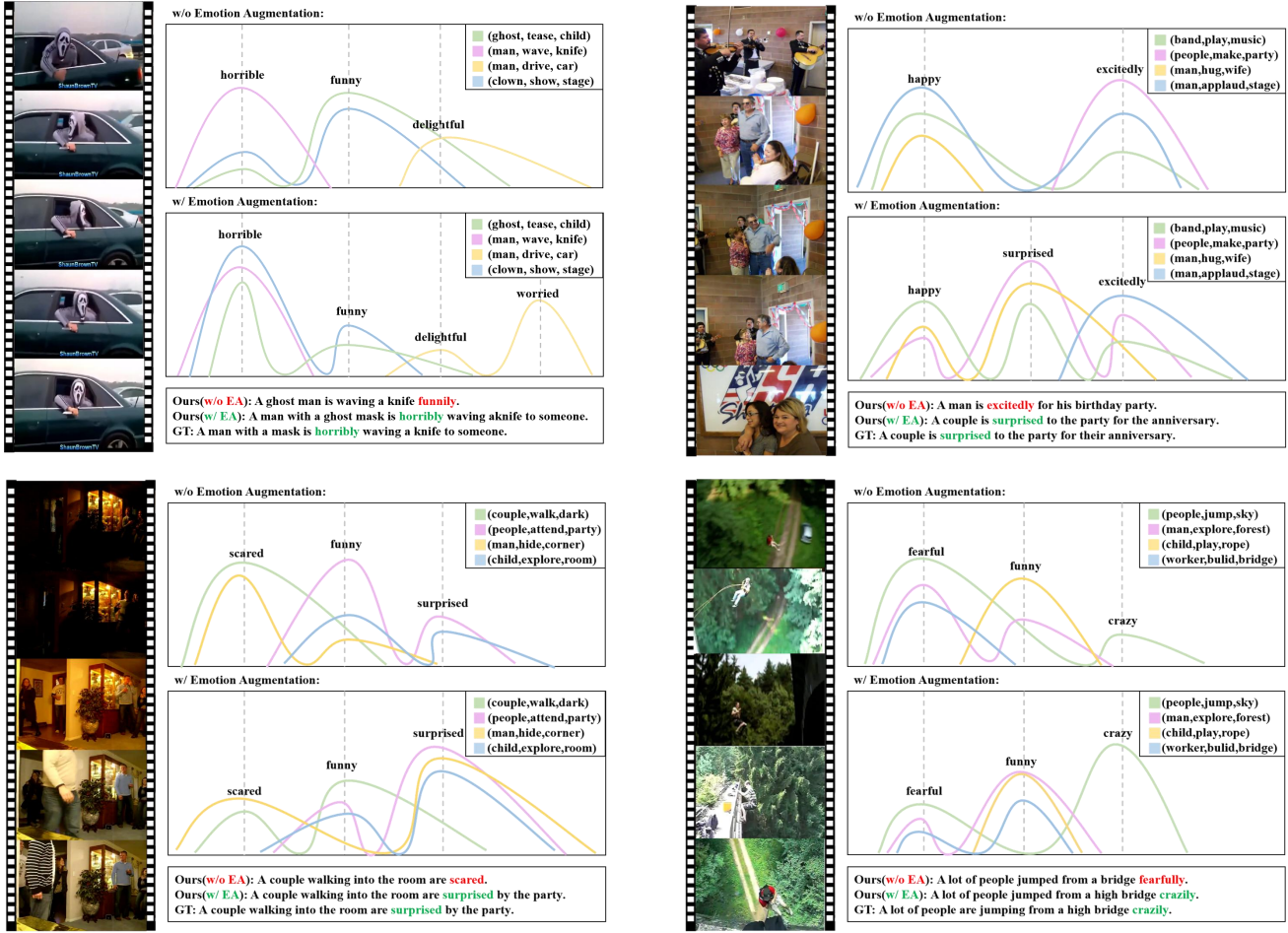

Fig. 8. Qualitative results to demonstrate the effectiveness of our proposed emotion augmentation module. The curve represents the emotional probability distribution under each triple.

### 4.5 Human Evaluation

EVC is essentially a text generation task. It is a crucial criterion whether the generated descriptions conform to human preferences for evaluating the model performance. Therefore, we add human evaluation to fully measure the quality of emotional descriptions. Inspired by the work [7], we designs four metrics, including (1) emotion accuracy: evaluate the accuracy of emotion in the sentence, (2) relevance: assess whether the sentence is relevant to the video content, (3) coherence: judge the logic and readability of the sentence, and (4) usability: how useful would the sentence be for a person (especially a blind person) to understand what is happening in the video. For each of these four metrics, the score ranges from 1 to 10; and the higher the score, the better the performance on the metric. Subsequently, we invite 10 participants with excellent English skills to score a subset of 50 video-caption pairs on these four metrics. For the state-of-the-art models and our model, we calculate the average of scores on these four metrics, respectively. As shown in Table 7, our model is significantly ahead of the state-of-the-art methods in all four metrics, which shows that our model does not simply imitate the reference sentences, but learns the representation of emotional and factual semantics.

### 4.6 Qualitative Results

To intuitively demonstrate the performance of our model, we present several qualitative results. Firstly, we make a visualization comparison with the SOTA methods. Secondly, we present two failure cases of our proposed method. Besides, we make a visualization for our proposed factual calibration and emotion augmentation module, respectively.

**Comparison with previous methods.** To qualitatively demonstrate the effectiveness of our method, we make a

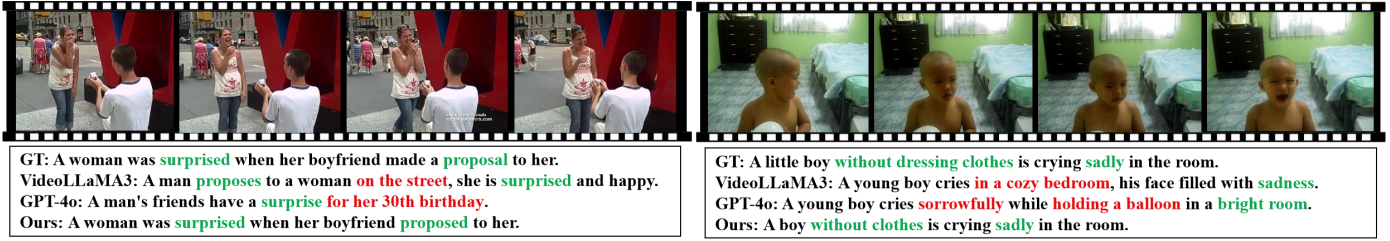

GT: A woman was **surprised** when her boyfriend made a **proposal** to her.
VideoLLaMA3: A man **proposes** to a woman **on the street**, she is **surprised** and happy.
GPT-4o: A man's friends have a **surprise for her 30th birthday**.
Ours: A woman was **surprised** when her boyfriend **proposed** to her.

GT: A little boy **without dressing clothes** is crying **sadly** in the room.
VideoLLaMA3: A young boy cries **in a cozy bedroom**, his face filled with **sadness**.
GPT-4o: A young boy cries **sorrowfully** while **holding a balloon** in a **green room**.
Ours: A boy **without clothes** is crying **sadly** in the room.

Fig. 9. Qualitative results to compare the effectiveness between our model and LVLMs. The red words indicate incorrect words generated by LVLMs due to hallucinations.

TABLE 8
The results of human evaluations of our model and LVLMs GPT-4o and VideoLLaMA3.

| Methods | Accuracy | Relevance | Coherence | Usability |
|---|---|---|---|---|
| GPT-4o | 6.98 | 6.04 | 7.24 | 6.62 |
| VideoLLaMA3 | 7.22 | 6.20 | 7.72 | 6.84 |
| **FACE-net(Ours)** | **7.62** | **7.24** | **8.50** | **8.06** |

visualization comparison with the state-of-the-art method DCGN. As shown in Fig. 4, our model could generate accurate and detailed descriptions. For example, in the first case, DCGN roughly describes the man's behavior as "*strange*", losing the detail of the video content and reducing factual descriptions. Instead, our module enhances factual descriptions by retrieving external repositories with calibration, which generates accurate descriptions "*lose his head*".

**Failure Cases.** Besides, two failure cases of the proposed method are shown in the last two samples of Fig. 4. In the first case, when processing videos with a large number of objects and complex relationships, the generated captions of the LLM-based decoder may produce hallucinations, significantly reducing the performance. In the second case, due to the limited size of the retrieval repository, the model may not be able to find content that is highly similar to the video in the retrieval repository. Thus, the trade-off between repository size and lightweight module design needs to be further researched. Meanwhile, the generalization capability is always a challenge for all vision-and-language model. In real-world applications, videos are more generalized, including various videos from different domains. Our method is still a certain distance away from practical applications in the real world scenarios. In the future work, we will connect our method to the existing large vision-language models, like GPT-4o, to obtain practical applications. Meanwhile, we will also continue to work on model lightweighting and generalization capabilities.

**Evaluation on Factual Calibration.** In addition, we make a qualitative comparison to show the effect of factual calibration module. As shown in Fig. 7, with the self-refinement component, our model could focus on the key word such as "tooth". Besides, we observe that the retrieved triplets (*child, play, game*) and (*boy, jump, rope*) contain irrelevant information. Without the cross-refinement component, our model incorrectly judges the object of the video content from "*tooth*" to "*rope*" and generates the wrong captions "*the girl plays her rope delightedly*". In contrast, with the cross-refinement component, our model could generate a low-weight refinement factor to reduce the impact of irrelevant information, *i.e.*, 11% for (*child, play, game*) and 4% for (*boy, jump, rope*), which leads to generate a more accurate caption.

**Evaluation on Emotion Augmentation.** Furthermore, we perform some cases to demonstrate the effect of emotion augmentation module. As shown in Fig. 8, we show the emotional distribution for each triplet with and without the emotion augmentation module. Without emotion augmentation, the triplets (*ghost, tease, child*) and (*clown, show, stage*) express the main emotion of "*funny*", which misleads the emotional direction and generates the wrong caption "*a ghost man is waving a knife funnily*". Instead, with emotion augmentation, our model achieves emotional alignment of triples and videos from "*funny*" to "*horrible*" by adaptive emotion mining, which leads to accurate emotion guidance.

### 4.7 Comparison with Large Vision Language Models (LVLMs)

We make a comparison between our model and existing LVLMs such as VideoLLaMA3[2] and GPT-4o[3]. Specifically, for testing, we provide the video to the LVLMs and give the following prompts:

*"I have given you the video. Please generate an English caption of no more than 15 words for this video. Please carefully analyze the visual content and emotional semantics in the video."*

We select the subset of EVC-MSVD as the test set. The human evaluation results are shown in Table 8 and the visualization results of some samples are shown in Fig. 9. We can observe that both GPT-4o and VideoLLaMA3 perform poorer than our model with simple prompts on the EVC task. Existing LVLMs are not sensitive to emotional perception, and cannot adaptively coordinate factual and emotional semantics to solve the F-E bias. Meanwhile, hallucinations are also an urgent dilemma for LVLMs, leading to poor performance on the EVC task.

### 5 Conclusion

In this paper, we first observe and define the factual-emotional bias problem, namely different samples in the dataset have different tendencies toward facts or emotions in EVC task, which is ignored by previous works. To this end, we propose a novel retrieval-enhanced framework with **FA**ctual **C**alibration and **E**motion augmentation (FACE-net) for EVC, which through a unified architecture collaborative mines factual-emotional semantics and provides adaptive and accurate guidance for caption generation. Our framework firstly introduces an external repository and retrieves the most relevant sentences with the video content to augment the semantic information. Subsequently, a factual calibration via uncertainty estimation module is proposed to effectively eliminate noise in retrieved information to calibrate and obtain the factual semantics by self-refinement and cross-refinement component, respectively. Besides, a progressive visual emotion augmentation module is proposed to leverage the calibrated factual semantics as experts, interact with both video contents and emotion dictionary to augment adaptive emotion semantics. Finally, we aggregate these multimodal semantic representations and leverage the powerful LLMs-based decoder to generate vivid emotional descriptions. Extensive experiments on three challenging datasets demonstrate the superiority of our method and each proposed module.

---

2. https://github.com/DAMO-NLP-SG/VideoLLaMA3
3. https://openai.com/index/hello-gpt-4o/

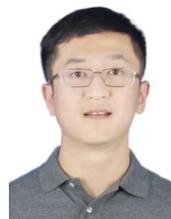

**Weidong Chen** (member, IEEE) received the Ph.D. degree in computer application technology from University of Chinese Academy of Sciences, in 2022. He is currently an Associate Researcher with the School of Information Science and Technology, University of Science and Technology of China, Hefei, China. He was a post-doctor with the School of Information Science and Technology, University of Science and Technology of China, from 2022 to 2024. His research interests include computer vision, natural language processing and cross-modal understanding.

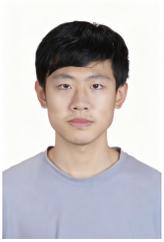

**Cheng Ye** received the B.E. degree in the School of Cyberspace Science and Technology, University of Science and Technology of China, Hefei, China, in 2024. He is currently a master student with the School of Information Science and Technology, University of Science and Technology of China, Hefei, China. His research interests include emotional intelligence, video analysis, and multimodal understanding.

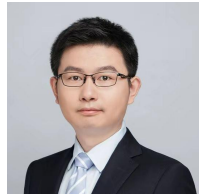

**Xiaojun Chang** (Senior Member, IEEE) is currently a Professor at the School of Information Science and Technology (USTC). He has spent most of his time working on exploring multiple signals (visual, acoustic, and textual) for automatic content analysis in unconstrained or surveillance videos. He has achieved top performances in various international competitions, such as TRECVID MED, TRECVID SIN, and TRECVID AVS.

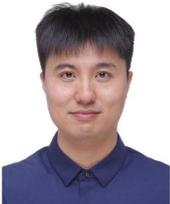

**Zhendong Mao** received the Ph.D. degree in computer application technology from the Institute of Computing Technology, Chinese Academy of Sciences, in 2014. He is currently a professor with the School of Cyberspace Science and Technology, University of Science and Technology of China, Hefei, China. He was an assistant professor with the Institute of Information Engineering, Chinese Academy of Sciences, Beijing, from 2014 to 2018. His research interests include cross-modal understanding and cross-modal generation. He serves as an Associate Editor of the IEEE TRANSACTIONS ON CIRCUITS AND SYSTEMS FOR VIDEO TECHNOLOGY and IEEE TRANSACTIONS ON MULTIMEDIA.

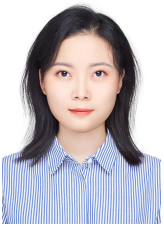

**Peipei Song** received the B.E. degree in electronic information engineering from Hefei University of Technology, China, in 2017, and the Ph.D. degree in signal and information processing from Hefei University of Technology, China, in 2023. She is currently a Associate Professor Fellow with the Department of Electronic Engineering and Information Science, University of Science and Technology of China (USTC). Her research interests include computer vision, multimedia analysis, and video understanding.

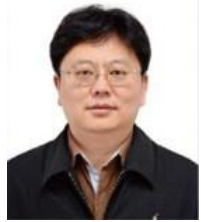

**Yongdong Zhang** (M'08–SM'13–F'24) received the Ph.D. degree in electronic engineering from Tianjin University, Tianjin, China, in 2002. He is currently a Professor with the School of Information Science and Technology, University of Science and Technology of China. His current research interests are in the fields of multimedia content analysis and understanding, multimedia content security, video encoding, and streaming media technology. He has authored over 200 refereed journal and conference papers, accumulating more than 29,000 citations on Google Scholar. He was a recipient of the best paper awards in PCM 2013, ICIMCS 2013, ICME 2010, the best student paper award in ACM Multimedia 2022 and the Best Paper Candidate in ICME 2011. He serves as an Editorial Board Member of the Multimedia Systems Journal and the IEEE TRANSACTIONS ON MULTIMEDIA. He is a fellow of the IEEE.

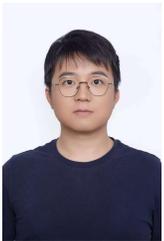

**Xinyan Liu** is currently an Associate Professor at Harbin Institute of Technology (HIT) and a Postdoctoral Fellow at City University of Hong Kong. He received his Ph.D. degree from the University of Chinese Academy of Sciences (UCAS) in 2025. His research interests lie at the intersection of video understanding and cross-media content understanding, with a focus on multimodal learning, video analysis, and intelligent content processing. He has published several academic papers in top-tier journals and conferences in her field. His current research explores the integration of computer vision and multimodal data analysis for advanced media intelligence applications.

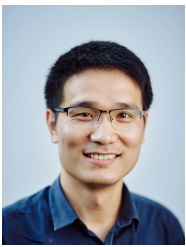

**Lei Zhang** received the Ph.D. degree in computer application technology from the Institute of Computing Technology, Chinese Academy of Sciences, in 2015. He is currently an associate professor with Department of Electronic Engineering and Information Science, University of Science and Technology of China, Hefei, China. Before that, he was a principal researcher at Kuaishou Technology. His research interests include large-scale retrieval, representation learning and cross-modal understanding.